\documentclass{article}
\usepackage{ijcai22}

\usepackage{times}
\usepackage{soul}
\usepackage{url}
\usepackage[hidelinks]{hyperref}
\usepackage[utf8]{inputenc}
\usepackage[small]{caption}
\usepackage{graphicx}
\usepackage{amsmath}
\usepackage{amsthm}
\usepackage{booktabs}
\usepackage{algorithm}
\usepackage{amsfonts}
\usepackage{amssymb}
\usepackage{algpseudocode}
\usepackage{color}
\usepackage{xspace}
\newcommand{\ie}{\emph{i.e.,}\xspace}
\newcommand{\eg}{\emph{e.g.,}\xspace}

\title{Cross-token Modeling with Conditional Computation}

\author{
    Yuxuan Lou,
    Fuzhao Xue,
    Zangwei Zheng,
    Yang You
    \affiliations
    National University of Singapore, Singapore
    \emails
    \{yuxuanlou, f.xue\}@u.nus.edu, zhengzangw@gmail.com, youy@comp.nus.edu.sg
}

\begin{document}

\maketitle

\begin{abstract}

Mixture-of-Experts (MoE), a conditional computation architecture, achieved promising performance by scaling local module (\ie feed-forward network) of transformer. However, scaling the cross-token module (\ie self-attention) is challenging due to the unstable training. This work proposes Sparse-MLP, an all-MLP model which applies sparsely-activated MLPs to cross-token modeling. Specifically, in each Sparse block of our all-MLP model, we apply two stages of MoE layers: one with MLP experts mixing information within channels along image patch dimension, the other with MLP experts mixing information within patches along the channel dimension. In addition, by proposing importance-score routing strategy for MoE and redesigning the image representation shape, we further improve our model's computational efficiency. Experimentally, we are more computation-efficient than Vision Transformers with comparable accuracy. Also, our models can outperform MLP-Mixer by 2.5\% on ImageNet Top-1 accuracy with fewer parameters and computational cost. On downstream tasks, \ie Cifar10 and Cifar100, our models can still achieve better performance than baselines.

\end{abstract}

\section{Introduction}

Mixture of Experts (MoE)~\cite{outrageously} is a powerful conditional computation architecture to scale the transformer up to trillions of parameters~\cite{switch_transformer}. However, although MoE can scale the local module (\ie feed-forward network) in transformer well, it is challenging to scale the cross-token module (\ie self-attention). The reason is that training a MoE-based attention is unstable and easy to diverge~\cite{switch_transformer}. In this work, we propose Sparse-MLP, a sparsely activated all-MLP model that can scale both local and cross-token module up efficiently.   


Sparse-MLP comprises a stack of Sparse blocks and Dense blocks. In each Sparse block, we apply sparsely-activated MoE layers at two stages. (1) Token-mixing MoE ($\mathrm{MoE_S}$), a computation-efficient alternative for self-attention, mixing the information across the spatial locations of image representations within channels. (2) Channel-mixing MoE ($\mathrm{MoE_C}$)), another stage of conditional computation, mixing the information within image representation patches along channels. In dense blocks, $\mathrm{MoE_S}$ and $\mathrm{MoE_C}$ are simplified as dense MLPs. Besides, we make two further improvements on MoE architecture. Firstly, we propose the importance-score routing strategy which can reduce routing computation of vanilla MoE. It ranks the tokens or channels for routing by their importance scores. Secondly, we redesign the image representation shape in Sparse blocks so that the gating network of token-mixing MoE can function with more efficiency.

A significant contribution of our work is, to our best knowledge, that we first set sparse Mixture of Experts as cross-token modeling module. The specific solution is using token-mixing MoE proposed to model the context. In general, we build our all-MLP model with conditional computation in two directions: both in patch dimension and channel dimension. It is also a major difference between our model and previous Transformer-MoE models~\cite{switch_transformer,vit_moe,widenet}. Previous models which apply MoE to transformer-based architecture only replace the FFN in the transformer block with sparse MoE. In our model, we have channel-mixing MoE layers function in a similar way, and token-mixing MoE layers function in another direction: mixing the information across the spatial locations of the representation. We prove with experiments that such a two-dimensional MoE design is effective and efficient.

Finally, We apply our Sparse-MLP models to image classification tasks and obtain outstanding results. After pre-trained with the self-supervised algorithm(MoCo v3)~\cite{moco} on ILSVRC2012 ImageNet-1k dataset~\cite{imagenet1k}, our Sparse-B model reaches $77.9\%$ ImageNet-1k top-1 accuracy, $2.0\%$ higher than Mixer-B/16 model with comparable computational cost, $1.2\%$ higher than ViT-B/16 with less computational cost. Our Sparse-L model reaches $79.2$ ImageNet-1k top-1 accuracy, outperforming Mixer-L/16 by $2.5\%$ with $62.8\%$ parameters and $85.5\%$ pre-training cost. It also outperforms ViT-L/16 by $1.8\%$ with less than half pretraining cost.

The contributions of this work can be summarized as follows:

\subsubsection{Two-stage MoE design and all-MLP architecture for cross-token modeling} We use computation-efficient MoE architecture as cross-token modeling module and then build an all-MLP architecture with two-stage MoE application. To our best knowledge, this is the first work focusing on scaling cross-token modeling module by conditional computation.

\subsubsection{Further efficiency improvement on MoE} We design the importance-score routing strategy which requires less computation cost without damaging model capacity. Also We revisit and redesign the image representation shape to make best use of token-mixing MoE. Both practice further improve our model's efficiency.

\subsubsection{Competitive performance on image classification tasks} We show that our Sparse-MLP model can outperform ViT model and dense MLP-Mixer model on Imagenet-1k benchmark. Also, on three downstream tasks, our Sparse-MLP models can reach better performance with comparable or less computational cost with same-level MLP-Mixer models.

\section{Cross-token Modeling with Conditional Computation}
The key contribution of our work is, we propose a sparsely-activated MLP architecture to efficiently and effectively for cross-token modeling: the token-mixing MoE layer. Token-mixing MoE follow the Mixture of Experts architecture \cite{outrageously}. For a batch of image inputs $X \in \mathbb{R}^{B \times S \times C}$ with batch size $B$, per image patches $S$ and per patch channels $C$, we firstly transpose inputs to $X' \in \mathbb{R}^{B \times C \times S}$. Then the gating network of token-mixing MoE assigns $B*C$ batch channels to different MLP experts by routing weights. Each expert model mixes the information along the patch dimension within each channel assigned to it. The details of token-mixing MoE layer and all-MLP architecture will be described in the following sections.

\subsection{Mixture-of-Experts}

In this section, we formulate Mixture-of-Experts (MoE) architecture and its key components.

\subsubsection{Conditional Computing}
The Mixture-of-Experts layer (MoE) is composed of a set of experts. Only a subset of them are active and engaged in the computation on a per-example basis. In our model, each expert is an MLP.

Following \cite{outrageously}, given $x \in \mathbb{R}^D$, the output of one MoE layer with $N$ Experts is:
\begin{equation}
\mathrm{MoE(x)}=\sum_{i=1}^{N} G(x)_i E_i(x)
\end{equation}
where $G(x): \mathbb{R}^D \rightarrow \mathbb{R}^N$ is the gating network which compute input-conditioned routing weights for experts. $E_i(x): \mathbb{R}^D \rightarrow \mathbb{R}^D$ is the $i^\mathrm{th}$ expert layer. In practice, we have a sparse $G(x)$, which means each input $x$ is restricted to be assigned to $k$ experts ($k \ll N$). If the input $x$ is not assigned to $E_i$, $G(x)_i=0$ and $E_i$ would not be computed. This enables us to scale to outrageously large model with a comparable computation cost. 

\subsubsection{Gating Network}
As we introduced above, to assign token representations $x$ to different experts, each MoE layer has a sparse gating network. We formulate it as: 
\begin{equation}
    G(x) = \mathrm{TopK}(\mathrm{softmax}(W_g(x)+\epsilon))
\end{equation}
where $W_g \in \mathbb{R}^{D\times N}$ is a trainable matrix and $\epsilon \sim  \mathcal{N}(0, \frac{1}{N^2})$ is a normal noise to explore better assignment from experts. After computing the probability of the input $x$ routed to each Expert, we only keep the top $K$ of them for further forward propagation. In practice, we usually select $K$ as 1 or 2.

\subsubsection{Load Balance Loss}
To encourage a balanced assignment of inputs across experts, an auxiliary loss is added to the model for every MoE layer \cite{outrageously,gshard,switch_transformer,vit_moe}. The formulation of load balance loss is in Appendix A.

\subsection{Importance-score Routing Strategy} \label{efficient}
As an attempt to further improve token-mixing MoE's efficiency, we propose the importance-score routing strategy. For each batch of image inputs after transpose $X' \in \mathbb{R}^{B \times C \times S}$, we set a importance score for all $B*C$ channels in the batch. Then, channels are sorted by their importance scores. $10\%$ with the lowest importance scores are eliminated and the rest are for the allocation.

In our work, the importance score of each channel for routing is its highest routing weight to all channels. Accordingly, $g(X)_{i,j}\in \mathbb{R}$ denotes the routing weight of i-th channel to the j-th expert. The importance score of i-th channel is:

\begin{equation}
    Score(\mathrm{Channel}_i)=max_j\{g(X)_{i,j}\}.    
\end{equation} 

In section \ref{abl_study}, we show empirically how such routing strategy can help further reduce computation cost without  damaging model capacity.

\begin{figure*}[t]
    \centering
    \includegraphics[width=0.9\textwidth]{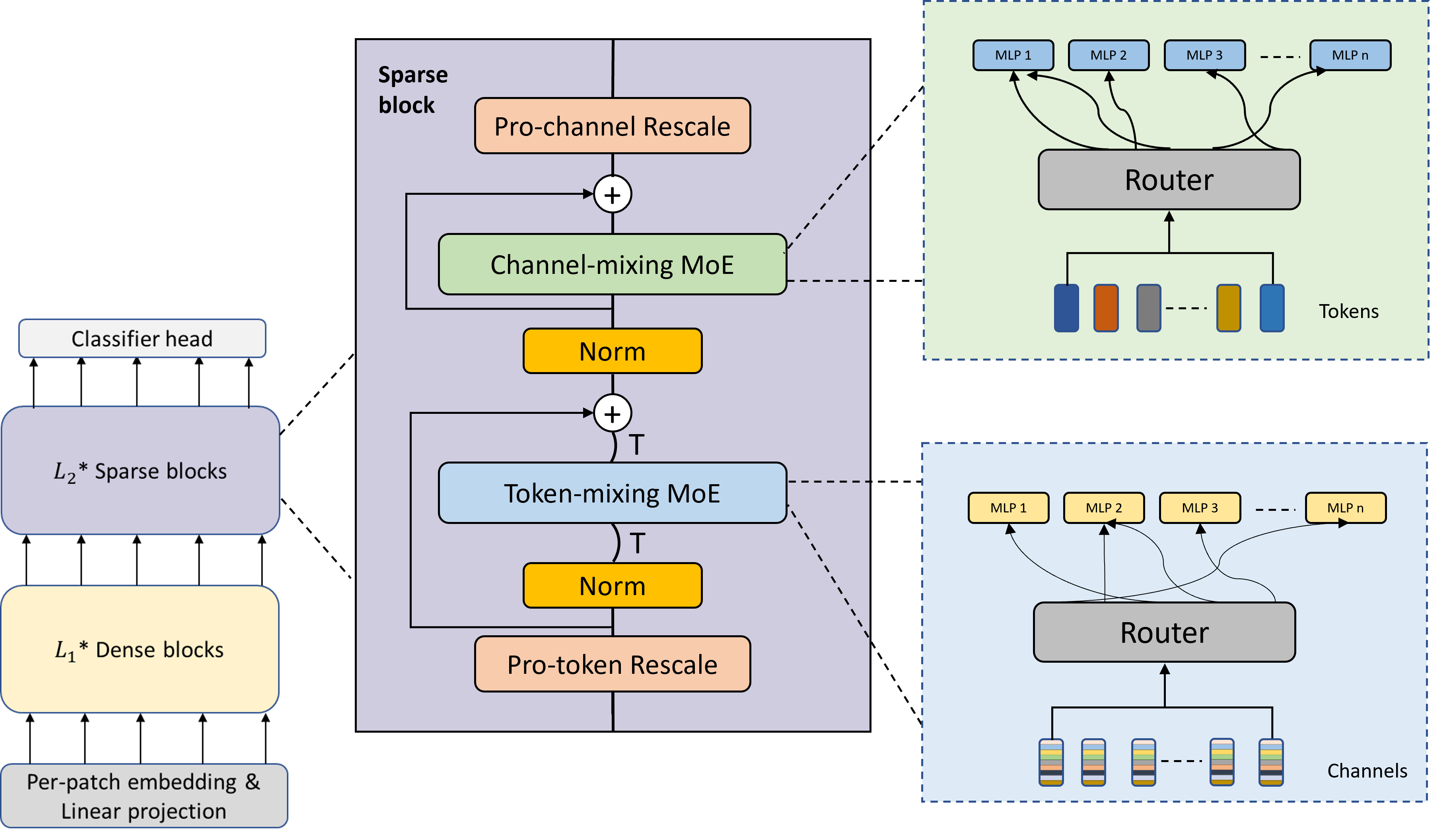}
    \caption{Sparse-MLP architecture overview}
    \label{fig:model_overview}
\end{figure*}

\subsection{All-MLP Architecture}
An overview of our full model, Sparse-MLP is shown in Figure~\ref{fig:model_overview}. In general, the Sparse-MLP model comprises a per-patch linear embedding layer, a stack of Dense blocks and Sparse blocks, and a classifier head. 

In each Sparse block, aside from the token-mixing MoE layer ($\mathrm{MoE_S}$) described above, we also apply a channel-mixing MoE layer($\mathrm{MoE_C}$), which is also a sparsely activated architecture and responsible for mixing the information along the channel dimension within patches. Besides, we have re-scale sub-layers at both the beginning and the end of each Sparse block which will be further described in Section \ref{rescale}. We formulate Sparse block as follows:

\begin{equation}
    x=\mathrm{Rescale}_1(x)
\end{equation}

\begin{equation}
    y_1=x + t(\mathrm{MoE_S}(t(\mathrm{norm}(x))))
\end{equation}

\begin{equation}
    y=y_1 + \mathrm{MoE_C}(\mathrm{norm}(y1))
\end{equation}

\begin{equation}    
    y=\mathrm{Rescale}_2(y)
\end{equation}

In each Dense block, both token-mixing MoE and channel-mixing MoE are simplified by single MLP: token-mixing MLP ($\mathrm{MoE_S}$) and channel mixing MLP ($\mathrm{MoE_S}$). The idea of Dense block follows \cite{mlp_mixer}. The formulation of Dense block is:

\begin{equation}
    y_1=x + t(\mathrm{MLP_S}(t(\mathrm{norm}(x))))
\end{equation}
\begin{equation}
    y=y_1 + \mathrm{MLP_C}(\mathrm{norm}(y1))
\end{equation}

In summary, our Sparse-MLP is an all-MLP model. And by applying sparse MoE architecture at two stages of information mixing, our model can achieve better model performance than other MLP-like models.

\section{Revisit Image Representation Shape} \label{rescale}
So far, we have an all-MLP model including token-mixing MoE and channel-mixing MoE. However, In the original ViT design \cite{vit}, the number of tokens is small while the per-token hidden size (\ie channel dimension $C$) is much larger. For image representation $x\in \mathbb{R}^{S \times C}, C\geq 3S$. If we adopt the original representation setting of ViT at token-mixing MoE layer, assignment of large amounts of channels to experts requires much computation while experts' capability are limited by short patch length. For example, in standard ViT setting, for a batch of input $X \in \mathbb{R}^{B\times S\times C}$, $B=4096,S=196,C=768$. The token-mixing MoE layer need to assign $B\times C=3145728$ channels to corresponding experts while experts' capability of mixing information within the channels is limited because the small output hidden size $S=196$.

In order to reduce the computation in assignment stage and improve token-mixing MoE's capability to mix information within channels, we add two stages of rescale sub-layers at the beginning and the end of each Sparse block. Pro-token rescale sub-layer, at the begining of each Sparse block, reduces number of channels and increases number of tokens so that the inputs can better fitting $\mathrm{MoE_S}$. Pro-channel rescale sub-layer, at the end of each Sparse block, functions the opposite way so that Sparse blocks and Dense blocks can be combined flexibly. 

Given an input $x \in \mathbb{R}^{S \times C}$, Pro-token rescale sub-layer maps:$\mathbb{R}^{C \times S} \rightarrow \mathbb{R}^{C_1 \times S_1}$. Pro-channel rescale sub-layer maps: $\mathbb{R}^{S_1 \times C_1} \rightarrow \mathbb{R}^{S \times C}$. Each rescale sub-layer is composed of two linear layers to transform two dimensions $S$ and $S_1$, $C$ and $C_1$, respectively. Such implementation can reduce routing computation and improve expert dimension, which leads to a more balanced and effective MoE design. In practice, we set $S_1 = 2S, C_1 = C/2$.
\section{Experiments}
We pretrain our Sparse-MLP models with MoCo V3 on the ILSVRC-2012 Imagenet dataset \cite{imagenet1k} and evaluate our model's performance on several downstream image classification tasks. We select MLP-Mixer models\cite{mlp_mixer} and ViT models \cite{vit} as our baselines and compare our models with baseline models in two quantities: (1) classification accuracy on downstream tasks, (2) computational cost of pre-training on the upstream dataset, and fine-tuning on downstream datasets. We do not aim to reach SOTA image classification accuracy but to show that our fully-MLP model with conditional computing can outperform dense MLP models or attention-based models either in accuracy or computational cost.

\subsection{Experiment Settings}

We pretrain Sparse-MLP models and baseline models(ViT and MLP-Mixer) with a self-supervised learning algorithm(MoCo v3) \cite{moco} on ILSVRC-2012 ImageNet dataset. \cite{imagenet1k}(1.3M training samples, 1k image classes) on TPU clusters. 

After pretraining, We fine-tune our model on three downstream tasks: ILSVRC-2012 Imagenet, CIFAR-10 (50k training samples, 10k validation samples, 10 classes) \cite{cifar10} and CIFAR-100. The detail setting of pretraining and fine-tune stage can be found in Appendix B.

\begin{table*}[htb]
    \centering
    \small
    \begin{tabular}{c|cccc}
    \toprule
         Models& ImageNet Top-1(\%) & Params(M) & Pre-training cost & Throughput\\ \midrule
         & \multicolumn{4}{c}{attention-based}\\ \midrule
         ViT-B/16 & 76.7& 86& 67.2&861\\
         ViT-L/16 & 77.6& 304&195.2&268\\ \midrule
         & \multicolumn{4}{c}{dense MLP-like} \\ \midrule
         Mixer-S/16 & 70.2& 19& 35.5& 3986\\
         Mixer-B/16 & 75.9& 59&53.3& 1320 \\
         Mixer-L/16 & 76.7& 207& 97.7& 412\\ \midrule
         & \multicolumn{4}{c}{Sparse-MLP}\\ \midrule
         Sparse-S & 71.3& 21& 35.5& 3986\\
         Sparse-B & 77.9& 69& 55.5& 1265\\
         Sparse-L & 79.4& 130 &80.1& 482\\ \bottomrule
    \end{tabular}
    \caption{ImageNet-1k results. All models are pretrained with self-supervised algorithm(MoCo v3) on ImageNet-1k and then fine-tuned. Pretrain cost is evaluated by total TPU v3 core-days used for pretraining. Throughput is evaluated by image/sec/core}
    \label{tab:imagenet_result}
\end{table*}

\subsection{Main Results} \label{main_results}

We build Sparse-MLP models on three parameter levels in comparison with attention-based models (\eg ViT~\cite{vit} ) and dense MLP models (\eg MLP-Mixer~\cite{mlp_mixer}). The specifications of our models can be found in Appendix C. In Table \ref{tab:imagenet_result}, we report ImageNet-1k top-1 accuracy and corresponding pre-training cost of each model. 

Our Sparse-S model surpasses Mixer-S/16 on ImageNet-1k top-1 accuracy by 1.1\% with comparable parameters and pre-training cost. Sparse-B model scales Mixer-B/16 with $17\%$ (59M$\rightarrow$69M) with comparable pre-training TPU v3 core days and outperforms Mixer-B/16 by $2.6\%$ ($75.9\% \rightarrow 77.9\%$). Our Sparse-L outperforms Mixer-L/16 by 3.3\% with only 62.8\% parameters and 85.5\% pre-training time. Compared with ViT, our models show better performance with much fewer parameters and much less pre-training cost.

\begin{figure}
    \centering
    \includegraphics[width=0.45\textwidth]{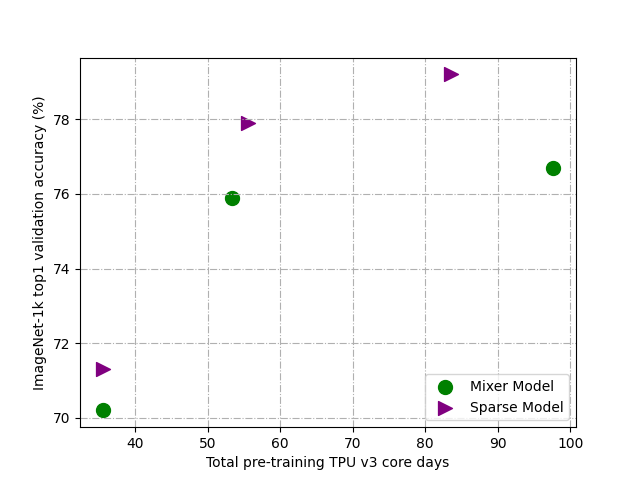}
    \caption{Comparison between Mixer models and Sparse-MLP models. With comparable or less computational cost, Sparse-MLP achieves better performance}
    \label{fig:main_results}
\end{figure}

\begin{table}[htb]
    \centering
    \begin{tabular}{c|ccc}
    \toprule
         Models& ImageNet & Cifar10 & Cifar100 \\
         & top-1 & top-1 & top-1 \\
         \midrule
         Mixer-S/16&70.2 & 91.7 & 84.4\\
         Sparse-S & 71.3& 91.9 & 84.4\\
         \midrule
         Mixer-B/16 &75.9 & 95.6&  86.7\\
         Sparse-B & 77.9& \textbf{96.2} & 87.2\\
         \midrule
         Mixer-L/16 & 76.7& 94.7& 86.3\\
         Sparse-L & \textbf{79.4}& 95.4 & \textbf{87.4}\\ \bottomrule
         
    \end{tabular}
    \caption{Results on downstream image classification tasks}
    \label{tab:downstream_results}
\end{table}

Also, we report the results of Sparse-MLP models and dense MLP models~\cite{mlp_mixer} on two other downstream image classification tasks: Cifar10 ~\cite{cifar10}, Cifar100~\cite{cifar}. All models are pretrained with MoCo v3 on ImageNet-1k and then fine-tuned at downstream tasks end-to-end.

In Table \ref{tab:downstream_results}, we can see that our Sparse-MLP models also outperform MLP-Mixer models on Cifar10 and Cifar100 image classification tasks. Also, when we scale our model to over 100M parameters, the performance of Mixer-L/16 and Sparse-L drop due to overfitting. This issue is prominent when training large MLP models on small datasets. And in such cases, our Sparse-L model still achieves higher accuracy than Mixer-L/16.

\subsection{Ablation Study} \label{abl_study}
In this section, we further investigate how each component of our Sparse-MLP model contributes to the performance. All models in the ablation study are pretrained with MoCo v3 algorithm on ImageNet-1k and fine-tuned on the same dataset. We select ImageNet-1k top-1 validation accuracy and total pre-training TPU v3 core days as evaluation metrics. The ablation study is designed to answer the following questions:
\begin{itemize}
    
    \item \textbf{Number of experts}: What is the impact of the number of experts in two stages MoE layers?
    \item \textbf{Top K routing}: Which K value(1 or 2) shall we select for $\mathrm{MoE_S}$ and $\mathrm{MoE_C}$?
    \item \textbf{Importance-score routing}: How much computation cost importance-score routing can save?
    \item \textbf{Positions of Sparse blocks}: How shall we combine Dense blocks and Sparse blocks?
    \item \textbf{rescale sub-layers analysis}: How do rescale sub-layers influence model performance and computational cost?
\end{itemize}

\begin{figure}[ht]
    \centering
    \includegraphics[width=0.4\textwidth]{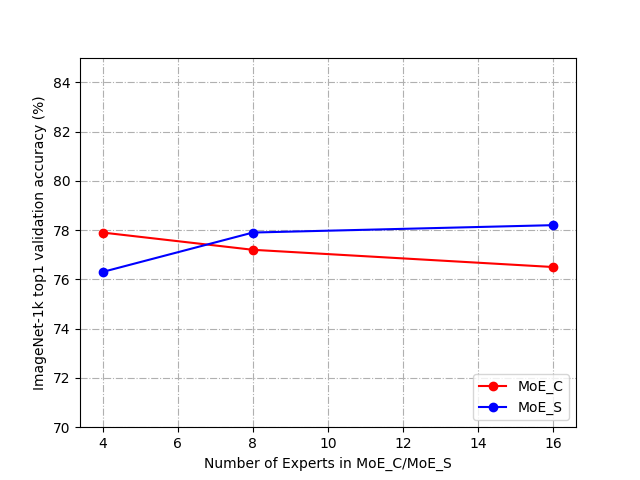}
    \caption{Influence of number of experts in $\mathrm{MoE_C}$/ $\mathrm{MoE_S}$}
    \label{fig:abl_moe}
\end{figure}

\subsubsection{Number of experts}
We first study the influence of the number of experts in $\mathrm{MoE_S}$ on model capacity. Different models are built based on Sparse-B. We fix all other hyper-parameters and tune the number of experts in $\mathrm{MoE_S}$ at three levels: 4, 8, 16, pretrain these models and evaluate their performance on ImageNet-1k validation top-1 accuracy.

From Figure~\ref{fig:abl_moe}, we can see that when the number of experts in $\mathrm{MoE_S}$ increases from 4 to 8, the model's performance increases a lot. However, when we scale experts to 16, the accuracy barely changes.

Similarly, for $\mathrm{MoE_C}$, we fix all other components in Sparse-B and tune the number of experts in $\mathrm{MoE_C}$ at three levels: 4,8, 16.

In Figure~\ref{fig:abl_moe}, we observe that there would be an overfitting issue when we increase the number of experts in $\mathrm{MoE_C}$. Such finding is similar to results in \cite{widenet}. When training data is limited, scaling the MoE layers, which mixes the information within spatial locations, will make the model easily overfit the training data.

\begin{table}[ht]
    \centering
    \small
    \begin{tabular}{c|cc}
         K& ImageNet Top-1(\%) & TPU v3 core days\\ \midrule
         & \multicolumn{2}{c}{$\mathrm{MoE_S}$}\\ \midrule
         1& \textbf{77.9}& \textbf{55.5}\\
         2 & 77.9&57.7\\ \midrule
         & \multicolumn{2}{c}{$\mathrm{MoE_C}$}\\ \midrule
         1 & 77.0& 53.3\\
         2 & \textbf{77.9}& \textbf{55.5}\\ \bottomrule
    \end{tabular}
    \caption{Influence of K selecting.}
    \label{tab:abl_k}
\end{table}
\subsubsection{The role of Top K routing}
Following the design in ~\cite{switch_transformer}, we select K=1 or 2 for $\mathrm{MoE_S}$ and $\mathrm{MoE_C}$. We set Sparse-B as our default model(K=1 for $\mathrm{MoE_S}$, K=2 for $\mathrm{MoE_C}$). Then we report the results with K=2 for $\mathrm{MoE_S}$ and K=1 for $\mathrm{MoE_C}$ separately. 

As shown in Table \ref{tab:abl_k}, for $\mathrm{MoE_S}$, top-1 routing and top-2 routing reach the same validation accuracy and top-1 routing cost less pre-training time. For $\mathrm{MoE_C}$, top-2 routing would lead to prominent better performance with $4\%$ more pre-training time.  

\begin{table}[ht]
    \centering
    \small
    \begin{tabular}{c|cc}
         K& ImageNet Top-1(\%) & TPU v3 core days\\ \midrule
         & \multicolumn{2}{c}{Sparse-B}\\ \midrule
         Efficient& 77.9& \textbf{54.3}\\
         Vanilla & 77.9&55.5\\ \midrule
         & \multicolumn{2}{c}{Sparse-L}\\ \midrule
         Efficient & \textbf{79.4} & \textbf{80.0}\\
         Vanilla & 79.2 & 83.5\\ \bottomrule
    \end{tabular}
    \caption{Importance-score Routing Versus Vanilla Routing}
    \label{tab:efficient_routing}
\end{table}

\subsubsection{Importance-score Routing Versus Vanilla MoE Routing}
In section \ref{efficient}, we propose a new routing method that requires less computation cost than vanilla MoE routing. We compare the two different routing methods on Sparse-B and Sparse-L models, we can see that with comparable results on ImageNet top-1 validation accuracy, importance-score routing strategy can reduce total pretraining time by 1.9\% and 4.1\%
\subsubsection{The positions of Sparse blocks}
We experiment with two different placing orders of Dense blocks and Sparse blocks. (1) Dense blocks in front and Sparse blocks behind; (2) Sparse blocks as first few blocks and followed by Dense blocks. Also, we experiment with a different number of Sparse blocks while keeping the number of total Dense blocks and Sparse blocks the same. We set Sparse-B as our default model and change the orders and numbers of blocks based on Sparse-B
\begin{table}[ht]
    \centering
    \small
    \begin{tabular}{c|cc}
    \toprule
    Positions& ImageNet Top-1(\%) & Parameters(M)\\
    \midrule
    N/A (Mixer-B/16) & 75.9 & 59\\ 
    Last two(Sparse-B)& 77.9& 69\\
    First two & 75.5& 69\\
    Last four & 78.3& 79\\ \bottomrule
    \end{tabular}
    \caption{Different combinations of Dense blocks and Sparse blocks. 'Positions' refers to the locations of Sparse blocks.}\label{tab:abl_position}
\end{table}

In Table \ref{tab:abl_position}, we can find that placing Sparse blocks in the first place and Dense blocks afterwards do not show good performance.
We also find that increasing the number of Sparse blocks, in the end, is an effective way to improve the model's performance. When we increase 2 Sparse blocks and keep the total number of blocks unchanged, the model's ImageNet-1k top-1 validation accuracy increased by $0.3\%$

\subsubsection{The role of rescale sub-layers}

An intuitive way to build Sparse blocks is to only apply two stages of Mixture of Experts with original image representation shape. As stated in Section \ref{rescale}, such design would lead to huge computational cost in routing stage of token-mixing MoE. Thus, we add rescale sub-layers to reduce computation cost of Sparse blocks. Here we verify the necessity of rescale sub-layers by experiments.
We set Sparse-B as our default model and experiment models with or without rescale sub-layers.

\begin{table}[ht]
    \centering
    \small
    \begin{tabular}{c|cc}
    \toprule
    Models& ImageNet Top-1(\%) & Pre-training cost\\
    \midrule
    w/ r\_layers&  \textbf{77.9}& \textbf{55.5}\\
     w/o r\_layers&  76.9& 79.9 \\ \bottomrule
    \end{tabular}
    \caption{Comparison between models with or without rescale sub-layers.}
    \label{tab:abl_rlayer}
\end{table}
We can see from table~\ref{tab:abl_rlayer} that rescale sub-layers not only reduce pre-training computation cost by $30.5\%$ but also improve the performance significantly.

\section{Why Token-mixing MoE and Sparse-MLP}
Although token-mixing MoE can be viewed as an computation-efficient alternative for self-attention, our Sparse-MLP model not only saves computational cost, but also have competitive performance. Sparse-MLP has its own advantage compared with ViT, previous transformer-MoE models and dense MLP-like models.
\begin{itemize}
    \item \textbf{Versus ViT}: Our model not only requires much less pretraining cost but also shows better model capacity empirically. In section \ref{main_results}, our model(Sparse-L) outperforms ViT at same parameter level with only less than half pretraining cost.
    \item \textbf{Versus previous transformer-MoE models}: Previous work applying MoE architecture to vision transformer models \cite{vit_moe} only replaces the FFN in the transformer block with Mixture of Experts layer. Like what attention does in transformer, we further apply the sparse MoE layer to cross-token dimension. In this way, we not only reduce computational cost, but also build an All-MLP architecture, which is more simple and easy to scale up.
    \item \textbf{Versus dense MLP-like models}: Compared with MLP-Mixer \cite{mlp_mixer}, our model scales up the single MLP to a set of sparsely-activated MLP structure. Such scaling strategy brings significant model performance improvements and only requires a sub-linear increase in computation cost. In section \ref{main_results}, Sparse-MLPs outperform Mixer models at same parameter level by a large margin.
\end{itemize}

\section{Related Work}
\subsection{Transformer-MoE models}
Mixture-of-Experts(MoE)~\cite{,outrageously} has recently been applied to transformer-based architecture to build huge models~\cite{gshard,switch_transformer,vit_moe,widenet}. In NLP tasks, ~\cite{gshard,switch_transformer} applied MoE to scale transformer-based models to trillions of parameters and achieve superising results. In vision tasks, ~\cite{vit_moe} improves ViT\cite{vit} by scaling a subset of transformer blocks with MoE. \cite{widenet} applies MoE to transformer blocks with parameters sharing to improve ViT with fewer parameters. In these works, MoE layers are to replace the FFN in transformer blocks. Our model design makes a difference in that we apply MoEs in two directions and experiments demonstrate that the novel token-mixing MoE can improve model capacity effectively and efficiently.

\subsection{MLP-like models}
Our work is also related to MLP-like models. Different from CNN models~\cite{alexnet,vgg} and attention-based models~\cite{vit,deit}, all trainable parameters in the backbones of MLP-based models are MLP-like. In MLP-Mixer~\cite{mlp_mixer}, a token-mixing MLP is to replace the multi-head self-attention~\cite{attention} in transformer block~\cite{mlp_mixer}. Some other MLP-like architectures~\cite{gmlp,permutator} function a similar way, mixing the information across spatial locations with MLPs or FFNs. Our  Sparse-MLP applies sparsely-activated MLP and can achieve better performance.

\section{Conclusion}
In this work, we propose token-mixing MoE, a sparse MoE architecture to model the cross-token information of images with conditional computation. Further, we propose Sparse-MLP, an all-MLP architecture with two-dimentional MoE in vision. Experiments demonstrate that our two-stage MoE with importance-score routing strategy and rescale sub-layer design are effective and computation efficient. Besides, we perform a comprehensive ablation study to investigate how each component contributes to the performance. 

Extensions of our work could include the following topics. First, it is possible to further improve Sparse-MLP model capacity with huge pre-training datasets. Besides, we can explore the flexibility of Sparse-MLP architecture by designing Sparse blocks with different MoE settings in the same model. It would also be worthwhile to apply Sparse-MLP architecture to NLP and other tasks.

\bibliographystyle{named}
\bibliography{ijcai22}

\clearpage
\appendix

\section{Load-balance loss for MoE} \label{load_loss}
Our auxiliary loss which encourages balanced routing consists of two parts: Importance loss and Load loss.

The importance of $i^\mathrm{th}$ expert is defined as the normalized gating network weights correspond to $i^\mathrm{th}$ expert summed over the input batch $X$. 
\begin{equation}
    \mathrm{Imp}_i(X)=\sum_{x \in X} \mathrm{softmax}(W_g x)_i
\end{equation}
where $W_g$ is the gating weight matrix of the MoE layer, and the importance loss of the MoE layer over a batch of inputs $X$ is:
\begin{equation}
    L_{imp}(X)=(\frac{\mathrm{std}(\mathrm{Imp}(X))}{\mathrm{mean}(\mathrm{Imp}(X))})^2
\end{equation}

In addition to the importance loss for more balanced routing weights, we also have a load loss seeking balanced routing results. The load of an Expert $i$ given a batch of inputs $X$ is defined as the possibility of routing to Expert $i$ summed over the batch.
\begin{equation}
    \mathrm{Load}_i(X)=\sum_{x \in X} p_i(x)
\end{equation}
\begin{equation}
    p_i(x) \triangleq P(G(x)_i)\geq threshold_k(G(x)))
\end{equation}
The load loss of one MoE layer over the batch is:
\begin{equation}
    L_{Load}(X)=(\frac{\mathrm{std}(\mathrm{Load}(X))}{\mathrm{mean}(\mathrm{Load}(X))})^2
\end{equation}
 And the total auxiliary loss of one MoE layer takes the form:
 \begin{equation}
     L_{aux} = \lambda(\frac{1}{2}L_{imp}+\frac{1}{2}L_{load})
 \end{equation}
where $\lambda$ is a hyper-parameter that controls that the auxiliary loss not only encourages balanced routing across experts but also not overwhelms the original model loss. In practice, we set $\lambda=1e-2$. According to existing MoE-based models~\cite{vit_moe,widenet}, the performance is insensitive to $\lambda$.

\section{Pretrain and fine-tune Details} \label{pretrain}
our data augmentation policy for pretraining includes random resized crop, horizontal flipping, RandAugment,  color jittering, grayscale conversion, blurring, and solarization. We also apply stochastic depth.

We pretrain all models on TPU v3 clusters. We select a batch size as 4096 at the pre-training stage, LAMB optimizer with weight decay. We pretrain all models for 300 epochs using a cosine learning rate decay with a 10k steps warm up. The image resolution for pretraining is 224.

At fine-tune stage, We follow the standard fine-tune settings in. After pretraining, we remove the MLP heads of the pretrained model, add a classifier head to the encoder, and train on downstream tasks. The augmentation strategies during fine-tuning stage include random resized crop, horizontal flipping, RandAugment  and Mixup. We select Adam without weight decay as the optimizer. We set our learning rate as $lr*\mathrm{Batch Size}/256$, using linear weight decay with 10k warm-up steps. Image resolution is $224 \times 224$.  
\begin{table}[ht]
    \centering
    \begin{tabular}{c|c}
    \toprule
         \textbf{Hyper-parameter} & \textbf{Value} \\
         \midrule
         Image resolution & 224\\
         Epochs & 300\\
         Batch size & 4096 \\
         Warmup steps & 10k \\
         Optimizer & LAMB \\
         Peak learning rate & 1e-3 \\
         Learning rate decay & cosine \\
         Weight decay rate & 1e-1 \\
         Global clip norm & 1 \\
         \midrule
          MoCo $t$ & 1\\
          MoCo $m$ & 0.99\\
          MoCo dim & 4096 \\ \bottomrule
         
    \end{tabular}
    \caption{Hyper-parameters for pre-training on ImageNet-1k}
    \label{tab:hp_pretrain}
\end{table}

\section{Model Settings} \label{model_setting}
We report our main results based on three models: Sparse-S, Sparse-B, Sparse-L. In Table \ref{tab:model_detail}, we give the specifications of these models. Each model is composed of $L_1$ Dense blocks and $L_2$ Sparse blocks. And in all three models reported in the main results, Dense blocks are in the front and followed by Sparse blocks. $D_S$ refers to the hidden dimension of token mixing MLPs, and $D_C$ refers to the hidden dimension of channel mixing MLPs. $D_{S'}$ is the MLP dimension of token-mixing MoE layers, and $D_{C'}$ denotes the MLP dimension of channel-mixing MoE layers. For all MLPs in Dense blocks and Sparse blocks, we set dropout as 0. For token mixing MoEs, we select top $K$ routing as 1. And for channel mixing MoEs, we set $K$ as 2.

\begin{table}[ht]
    \centering
    \small
    \begin{tabular}{c|ccc}
    \toprule
         Specification&Sparse-S &Sparse-B&Sparse-L \\
         \midrule
         &\multicolumn{3}{c}{Dense block}\\
         \midrule
         blocks $L_1$ & 6& 10& 8\\
         Patches $S$ &196 & 196& 196\\
         Hidden size $C$ & 512& 768&768\\
         $\mathrm{MLP_S}$ dim $D_S$ & 256& 384& 384\\
         $\mathrm{MLP_C}$ dim $D_C$ & 2048& 3072& 3072\\
         \midrule
         &\multicolumn{3}{c}{Sparse block}\\
         \midrule
         blocks $L_2$& 2& 2& 6\\
         New patches $S'$ & 392& 392&392\\
         New hidden size $C'$ &512 & 384& 384\\
         Experts in $\mathrm{MoE_S}$ & 4& 8& 16\\
         Experts in $\mathrm{MoE_C}$ & 0& 4& 4\\
         $\mathrm{MoE_S}$ top K & 1 & 1& 1\\
         $\mathrm{MoE_C}$ top K & - & 2& 2\\
         $\mathrm{MoE_S}$ dim $D_{S'}$ & 512& 768&768\\
         $\mathrm{MoE_C}$ dim $D_{C'}$ & 2048& 1536&1536\\

         Positions & last 2 & last 2 & last 6\\
         \midrule
         Parameters(M) & 22& 69&130\\ \bottomrule
    \end{tabular}
    \caption{Specifications of Sparse-MLP models}
    \label{tab:model_detail}
\end{table}

\begin{figure} 
    \centering
    \includegraphics[width=0.45\textwidth]{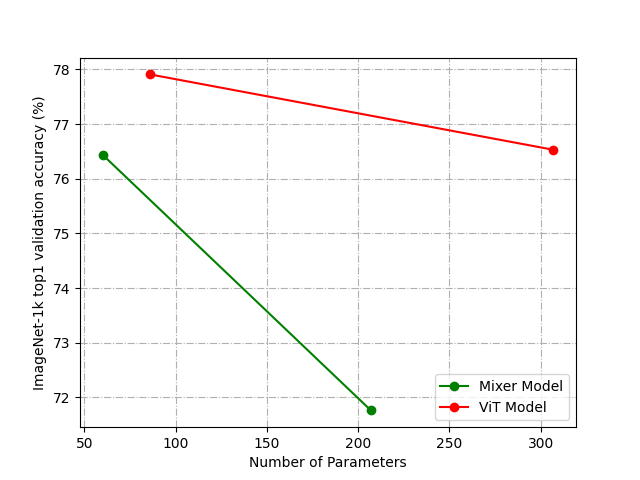}
    \caption{ImageNet-1k validation accuracy of ViT and MLP-Mixer, after supervised pretraining on ImageNet-1k}
    \label{appendix_pretrain}
\end{figure}

\section{Pretrain: Why Unsupervised}
We find that scaling MLP models or Vision Transformer models in parameters and training them from scratch with limited training data (\eg ImageNet-1k) will lead to an overfitting problem. As shown in Figure \ref{appendix_pretrain}, MLP-Mixer and ViT's accuracy both go down when parameters increase. Such finding is consistent with previous work on MLP models (\ie MLP-Mixer) and attention-based models (\ie ViT). 

In order to compare our models with baselines in a fairer way, and better evaluate models' performance when parameters scaling up, We adopt an unsupervised training algorithm: MoCo V3. We pretrain all our models on ImangeNet-1k dataset with MoCo V3 and then fine-tune them. Both Sparse-MLP and baseline models can achieve better performance with paramters increasing. 

\end{document}